\documentclass[12pt,draftclsnofoot,onecolumn,journal,letterpaper]{IEEEtran}

\usepackage[final]{graphicx}
\usepackage{amssymb}
\usepackage{url}
\usepackage{color}
\usepackage{times}
\usepackage{wrapfig}
\usepackage{subfig}
\usepackage{amsthm}
\usepackage{mathptmx}
\usepackage{amssymb}
\usepackage{amsfonts}
\usepackage{dsfont}
\usepackage{bm}
\usepackage{eucal}
\usepackage{mcite}
\usepackage{mathptmx}
\usepackage{multirow}
\usepackage{booktabs}
\usepackage{rotating}

\usepackage[cmex10]{amsmath}
\usepackage{amsfonts}
\def\mathbi#1{\textbf{\em #1}}
\usepackage{algorithm}
\usepackage{algorithmic}
\usepackage{array}
\usepackage{mdwmath}
\usepackage{mdwtab}
\usepackage{multirow}
\usepackage{fixltx2e}
\usepackage{cite}

\title{Designing Optimal Mortality Risk Prediction Scores that Preserve Clinical Knowledge}

\author{\IEEEauthorblockN{Natalia~M.~Arzeno~(1), Karla~A.~Lawson~(2), Sarah~V.~Duzinski~(2), Haris~Vikalo~(1)\\}
\IEEEauthorblockA{
(1) Department of Electrical and Computer Engineering,
The University of Texas at Austin, 
(2) Trauma Services, Dell Children's Medical Center of Central Texas \\
Email: narzeno@utexas.edu, hvikalo@ece.utexas.edu}}

\begin{document}

\maketitle

%%%%%%%%%%%%%%%%%%%%%%%%%%%%%%%%%%%%%%%%%%%%%%%%%%%%%%%%%%%%%%%%%%%%%%%%%%%%%%
%%%%%%%%%%%%%%%%%%%%%%%%%%%%%%%%%%%%%%%%%%%%%%%%%%%%%%%%%%%%%%%%%%%%%%%%%%%%%%

\begin{abstract}

Many in-hospital mortality risk prediction scores dichotomize predictive variables to simplify the score calculation. However, hard thresholding in these additive stepwise scores of the form ``add x points if variable v is above/below threshold t" may lead to critical failures. In this paper, we seek to develop risk prediction scores that preserve clinical knowledge embedded in features and structure of the 
existing additive stepwise scores while addressing limitations caused by variable dichotomization. To this end, we propose a novel
score structure that relies on a transformation of predictive variables by means of nonlinear logistic functions facilitating smooth 
differentiation between critical and normal values of the variables. We develop an optimization framework for inferring parameters of the 
logistic functions for a given 
patient population via cyclic block coordinate descent. The parameters may readily be updated as the patient population and 
standards of care evolve. We tested the proposed methodology on two populations: (1) brain trauma patients admitted to the intensive 
care unit of the Dell Children's Medical Center of Central Texas between 2007 and 2012, and (2) adult ICU patient data from the MIMIC 
II database. The results are compared with
those obtained by the widely used PRISM III and SOFA scores. The prediction power of a score is evaluated using area under ROC 
curve, Youden's index, and precision-recall balance in a cross-validation study. The results demonstrate that the new 
framework enables significant performance improvements over PRISM III and SOFA in terms of all three criteria.

\end{abstract}

%%%%%%%%%%%%%%%%%%%%%%%%%%%%%%%%%%%%%%%%%%%%%%%%%%%%%%%%%%%%%%%%%%%%%%%%%%%%%%
%%%%%%%%%%%%%%%%%%%%%%%%%%%%%%%%%%%%%%%%%%%%%%%%%%%%%%%%%%%%%%%%%%%%%%%%%%%%%%
\newpage
\section{Introduction}
\label{intro}
Technological advancements in medical instrumentation 
and a growing use of electronic medical records have created an abundance 
of clinical patient data. Extracting and analyzing useful information from such 
large and diverse data sets will enable tremendous advancements in clinical 
decision-making, ultimately leading towards improvements in health and 
quality of life as well as to reduction of the overall healthcare costs. Availability of
data has enabled development of accurate mortality and morbidity risk 
prediction scores for specific patient populations. Rapid prediction of potentially 
poor outcomes may provide timely intervention to reduce morbidity and mortality 
among the patients in the considered group.

Additive stepwise scores of the form ``add x points if variable v is above/ below threshold t" are popular tools for mortality risk prediction in both pediatric and adult intensive care unit (ICU) populations, typically using data acquired at the beginning of an ICU stay. Such scores comprise the Acute Physiology and Chronic Health Evaluation (APACHE) \cite{zimmerman_2006, knaus_1991}, the Simplified Acute Physiology Score (SAPS) \cite{moreno_2005, legall_1993}, and the Pediatric Risk of Mortality \cite{pollack_1996, pollack_1988}. However, while simple enough to allow for fast manual evaluation, these scores dichotomize
predictive variables to form prediction scores. Dichotomization of continuous variables 
results in a loss of information, increased probability of false negatives, and high 
dependence on cut-off points  \cite{streiner_2002,royston_2006}. This, in turn, may 
lead to critical failures of the prediction process. To remain relevant, risk scores need to be validated and updated periodically so that they reflect innovations and the evolution of the standards of healthcare \cite{marcin_2000, moreno_2005}, otherwise risking
deteriorating accuracy due to the changes in patient populations \cite{minne_2012}.
Moreover, 
risk scores might have higher accuracy for certain diseases \cite{zimmerman_2006} or may need 
to be customized for a subpopulation or a location \cite{moreno_2005}.

In this paper, we seek to develop risk prediction scores that preserve the clinical knowledge 
embedded in the features and structure of the aforementioned additive stepwise risk scores 
while allowing for soft thresholds in the score calculations. This is accomplished by a novel
scoring mechanism that relies on a transformation of predictive variables by means of nonlinear 
logistic functions to facilitate smooth differentiation between critical and normal values of the 
variables. The parameters in the proposed framework can be readily optimized for a specific 
sub-populations of interest (e.g., particular disease or location) or re-learned to ensure that the 
score remains relevant as the standards of care evolve. We use the PRISM III score 
\cite{pollack_1996} and a pediatric brain trauma population as well as the Sequential 
Organ Failure Assessment (SOFA) score \cite{vincent_1996, vincent_2000} and an adult ICU 
population, as motivation and to test the performance of the novel scoring mechanism. The 
paper is organized as follows. Some of the most common risk scores used in the pediatric and 
adult ICU are reviewed in Section \ref{sec:relWork}. In Section \ref{sec:methods}, 
we present the novel scoring mechanism, describe an algorithm for finding the optimal 
parameters of the logistic functions used to transform the data, and show quasiconvexity of the 
corresponding optimization problem. Section \ref{sec:results} presents the results obtained 
by applying the proposed methodology to predict mortality of pediatric trauma patients admitted 
to the ICU of the Dell Children's Medical Center of Central Texas between 2007 and 2012 (Subsection \ref{sec:resultsPedi}) and predict mortality of adult ICU patients from the MIMIC database (Subsection \ref{sec:resultsAdult}). Section \ref{sec:conclusion} 
concludes the paper.

\section{Existing Risk Scores} \label{sec:relWork}
In this section we overview existing prognostic models for both pediatric and adult 
patient populations, with the emphasis on PRISM III and SOFA. We also briefly mention data-driven 
techniques that have recently been used in model development and close the section by 
describing recent publications that incorporate expert knowledge in their models.

\subsection{Pediatric population risk scores}
An example of the pediatric risk prediction scores, the Pediatric Risk of Mortality (PRISM III) 
\cite{pollack_1996}, is a widely used scoring mechanism in pediatric ICU (PICU) \cite{marcin_2000} 
that has been validated in various settings as both an individual predictor and 
a significant predictor in a multivariate model in the United States and internationally 
\cite{scavarda_2010, gemke_2002, brady_2006, karambelkar_2012, cantais_2001, 
bahloul_2011, volakli_2012, martha_2005, qureshi_2007}. A state-of-the-art scheme, 
PRISM III has an additive stepwise structure 
that relies on $17$ physiological variables and $26$ ranges. The physiological variables it 
considers are characterized by their maximum or minimum values recorded during the first 
$12$ or $24$ hours after a patient's admission to the PICU. These variables include the 
minimum systolic blood pressure, maximum heart rate, the presence of fixed pupils, maximum 
and minimum body temperature, and a variety of laboratory measures including minimum and 
maximum CO\textsubscript{2} and pH, white blood cell count, glucose and platelet count. 
Contribution of some variables to the score is determined after evaluating a judiciously chosen 
logical OR statement, such as in the case of maximum prothrombin time (PT) and partial 
thromboplastin time (PTT), which can both detect abnormalities in clotting time. Other variables, 
such as systolic blood pressure and heart rate, have age-dependent ranges. 

In PRISM III, the score is incremented when the maximum (minimum) value of a variable in the 
score is above (below) a predetermined threshold. For example, if a child has a minimum 
Glasgow Coma Scale (GCS) \cite{laurer_2002,maas_2011} score less than $8$, then $5$ points 
are added to her/his PRISM III score. Clearly, calculation of the score is highly dependent on the 
established cutoff points and thus the prediction may change abruptly due to very small changes 
in the underlying variables. For instance, interpretation of the variables such as heart rate or blood 
pressure, which are altered by the simple act of breathing, may widely change due to the strict 
threshold structure -- an adolescent with a maximum heart rate of $144$ beats per minute is 
considered healthy, while another one with a single measure of $145$ beats per minute has $3$ 
extra points added in the PRISM III score calculation. Although the PRISM scores have been 
validated in numerous settings, they have also been shown to overpredict \cite{slater_2004,tibby_2002} 
and underpredict \cite{bhadoria_2008,thukral_2006} PICU deaths. Poor patient discrimination by 
PRISM scores, especially in neonates and infants \cite{wells_1996,goddard_1992}, and the fact
that only a small subset of PRISM variables are significant predictors of the outcome \cite{ponce_2005},
render the PRISM scores sensitive to population characteristics and standards of care and
suggest that they may not necessarily be institution independent. Unlike in
PRISM III (and the previous versions of PRISM) where the score is computed based on binary indicators of the raw 
feature values, in this paper each feature is first transformed by a non-linear logistic function 
whose inflection point and slope we find via an optimization procedure. As a result, we identify a 
range of values for which the risk changes continuously and monotonically (i.e., increases or 
decreases with the feature values), which stands in contrast to describing the effect of physiological 
variables on the risk of mortality by comparing the variables to pre-defined thresholds (the
strategy employed by state-of-the-art prediction schemes such as PRISM III).

In addition to PRISM III, widely-used scores in the PICU include the injury severity score (ISS) 
\cite{baker_1974} and the pediatric index of mortality (PIM2) \cite{slater_2003}, where the former is 
specific to trauma patients. The ISS is an anatomic score based on the location and severity of the 
injuries. Limitations of the ISS have led to various modifications as well as risk scores that incorporate 
the ISS in the calculation \cite{marcin_2002}. In a pediatric trauma population, PRISM has 
outperformed the ISS and its variants in identifying in-hospital mortality \cite{cantais_2001}. Logistic 
regression has often been employed to learn the weighting coefficients for physiologic variables or 
binary indicators in mortality prediction models \cite{marcin_2002}. PIM2 is a second generation score, 
based on recalibrating coefficients of PIM \cite{shann_1997} and adding variables for diagnostic 
groups with poor performance or calibration. PIM2 models risk using logistic regression with $10$ 
variables acquired upon hospital admission or in the first hour after PICU admission, $7$ of which are 
binary indicators. Specifically, the continuous PIM2 variables include first systolic blood pressure, 
ratio of FiO\textsubscript{2} to PaO\textsubscript{2}, and absolute arterial or capillary base excess. 
Fixed pupils, mechanical ventilation, elective admission, PICU admission for procedure recovery, and
cardiac bypass are included in the model as binary variables. Finally, a selection of high risk or low 
risk diagnoses, where the model was found to over- or under-estimate mortality, complete the list of
logistic regression variables \cite{slater_2003}.

\subsection{Adult population risk scores}
The Sequential Organ Failure Assessment (SOFA) \cite{vincent_1996,vincent_2000} 
 is an additive stepwise score that assigns a score of 
0-4 to each of six physiological systems: respiratory, coagulation, hepatic, cardiovascular, 
renal, and central nervous systems. The total SOFA score is then evaluated as the sum 
of the individual system scores. SOFA was originally designed to describe the degree 
of organ dysfunction in patients rather than be a mortality prediction score. However, 
Vincent et al. acknowledge that there is a correlation between the mortality rate and 
organ dysfunction \cite{vincent_1996,vincent_2000}. Subsequently, SOFA has been 
successfully implemented as a mortality prediction score, achieving performance 
comparable to that of other established risk-prediction scores \cite{minne_2008, 
hwang_2012}. Studies developing SOFA-based models consider the score at either 
a fixed time or incorporate sequential measurements where the difference in SOFA 
might be indicative of the risk of mortality \cite{minne_2008}. Going a step beyond 
examining the delta SOFA values, Toma et al. developed risk prediction models that 
include temporal patterns from SOFA measurements \cite{toma_2007}. Moreover,
SOFA has been shown to improve the accuracy of mortality risk-prediction of 
established scores such as APACHE and SAPS when used in combination with 
them \cite{fueglistaler_2010, ho_2007}.

Commonly used mortality risk prediction scores developed for adult populations \cite{breslow_2012a} include the popular APACHE \cite{zimmerman_2006, knaus_1991}, SAPS \cite{moreno_2005, legall_1993}, and Mortality Prediction Model at ICU admission (MPM\textsubscript{0}) \cite{lemeshow_1993,higgins_2007}. SAPS 3 \cite{moreno_2005} and MPM\textsubscript{0}III \cite{higgins_2007} use data obtained within one hour of ICU admission, while APACHE IV \cite{zimmerman_2006} is calculated from data obtained in the first 24 hours of the ICU stay. MPM\textsubscript{0}III is calculated from 15 binary features and age, where the binary features comprise dichotomized physiological variables, CPR before admission, mechanical ventilation, and the existence of certain chronic and acute diagnoses. In addition to these features, MPM\textsubscript{0}III includes 7 two-way interaction terms. Because feature weights are coefficients from a logistic regression, the weights can be positive or negative. SAPS 3 also has either positive or negative contribution from its variables, but with integer-valued weights. Twenty variables are required for SAPS 3 calculation; these variables can be divided into 3 groups: information before admission such as comorbidities and medications, circumstances of ICU admission such as whether it is a planned admission, and physiological variables. Unlike MPM\textsubscript{0}III, variable contribution to SAPS 3 follows an additive stepwise structure with multiple hard thresholds for the physiological variables. Mortality risk based on SAPS 3 is additionally differentiated from other scores by the existence of multiple formulas to calculate risk based on the geographic region of the patient. Like SAPS 3, APACHE is an additive stepwise score. The most recent iteration, APACHE IV \cite{zimmerman_2006}, includes 142 variables, 115 of which are categorical for admission diagnoses. Some of the variables in the model, such as age and the acute physiology score (APS), have linear and nonlinear contribution in the form of restricted cubic splines. The APS is in turn calculated as the APACHE III \cite{knaus_1991} score in the first 24 hours of the ICU stay. The variables in APACHE III are divided into 3 groups: age, physiology, and chronic health. The chronic health variables specify different point contributions based on certain comorbidities, while age and the 17 physiological variables contribute to the score in an additive stepwise fashion. APACHE III includes interactions between variables in the form of AND statements, where the integer-valued contributions to the score are dependent on two physiologic variables being within specified ranges.  A common customization of prognostic models requires relearning the feature weights for a given population, as in the case of APACHE IV which has been successfully implemented to predict mortality at different time points in a Dutch ICU population \cite{brinkman_2013}. While many have focused on validating and comparing the prognostic models, as described in review articles \cite{keegan_2011,vincent_2010}, others have sought to understand which component of a score is most predictive of mortality \cite{knox_2014} or whether the addition of a particular variable such as age or resuscitation status will increase the discriminatory capability of a model \cite{knox_2014, keegan_2012}. However, these modifications do not address the variable dichotomization limitation of these scores.

\subsection{Data driven risk scores and the preservation of expert knowledge}
Going beyond the traditional statistical methods, data mining and machine learning techniques 
have recently been applied to the development of prognostic models seeking to aid in 
determining the time for treatment initiation, therapy choice, and healthcare quality assessment 
\cite{clinical}. Recently used methods include decision tree techniques 
\cite{courville_2009, gortzis_2008}, neural networks \cite{gortzis_2008}, topic models \cite{lehman_2012}, autoregressive implementation of PRISM \cite{ruttimann_1993}, and 
techniques that incorporate injury coding schemes in the models \cite{burd_2009}. 
While several scores with additive stepwise structure exist for adult populations (such as APACHE 
and SAPS), most pediatric 
mortality risk prediction scores assume that risk depends linearly on the variables and do not consider
nonlinear variable transformations. PRISM III is an exception since it uses thresholding to prevent 
uninformative increase in risk at very high or very low variable values. In neonate morbidity prediction, 
the PhysiScore \cite{saria_2010} achieves greater predictive power than previously established 
neonatal morbidity scoring systems by relying on a nonlinear transformation of the raw variables in 
the feature set. In particular, that work employs nonlinear Bayesian models based on log odds ratios 
of the risk derived from the probability distribution that provides the best fit to the data for each of two 
patient classes.

Additive stepwise scores, including PRISM III and SOFA, have a strong dependence on expert knowledge during the developmental stage. Paetz \cite{paetz_2004} proposed a data-driven method for designing additive stepwise scores, where the step weights and the variable ranges are randomly initialized and learned by means of evolutionary strategies. This method does not require experts until the final fine-tuning stage if such a stage is deemed necessary, which presents many advantages in the initial stages of the score design. However, training of the score does not allow for missing data, which is unlikely in the typical practical scenarios 
where many variables need to be collected and analyzed.

Models that integrate domain expert knowledge with a data driven approach have been reported to 
result in greater predictive accuracy. In \cite{sun_2012}, combining knowledge based and data driven 
risk factors in a prediction model for heart failure greatly improved on the performance of a solely 
knowledge based classifier while still resulting in a clinically meaningful model. In the task of identifying 
similar concept pairs in clinical notes, combining context-based similarity and knowledge-based similarity 
in an algorithm has likewise resulted in a more accurate similarity score \cite{pivovarov_2012}.

\section{Methods} \label{sec:methods}
Most of the existing scores described in Section \ref{sec:relWork} require clinical expert 
input in the score development stage to determine thresholds for physiological variables. Data-driven 
approaches largely ignore expert knowledge in order to achieve the most discriminative results for a
given dataset, which may lead to lack of interpretability. The method for designing an optimal risk 
prediction algorithm described in this section employs data-driven techniques while preserving the 
expert knowledge embedded in the existing risk scores. In particular, in this section we propose a 
novel outcome prediction score that relies on a nonlinear transformation of the features, present an 
algorithm for optimizing the parameters of the transformation, and discuss optimality of the 
aforementioned algorithm.

\subsection{The new score and an algorithm for optimizing parameters of the logistic transformation 
of predictive features}

We describe the risk of mortality using a logistic regression model, where the conditional probability 
that patient $i$ dies during the hospital stay is given by
\begin{equation} \mathbf{P}(y_i=1 | \mathbf{w}, \mathbf{z}_i) = \frac{1}{1+\exp(-\mathbf{w}^T\mathbf{z}_i)},
\end{equation} 
and the conditional probability of survival is 
\begin{equation} \mathbf{P}(y_i=-1 | \mathbf{w}, \mathbf{z}_i) = \frac{1}{1+\exp(\mathbf{w}^T\mathbf{z}_i)},
\end{equation} 
where $y_i \in \{-1,1\}$ is an indicator of the in-hospital mortality, the vector
$\mathbf{w}=[w_1 \; w_2 \; \ldots w_{M_w}]'$ collects weights
for the features $\mathbf{z}_i = [{z}_{i1} \; {z}_{i2} \; \dots \; {z}_{iM_w}]'$, and $M_w$ denotes the total
number of features. 

In a departure from the commonly used hard thresholding of predictive features and discrete scoring, we introduce a logistic transformation of the predictive features. The resulting new score is continuous and differentiable which enables computationally efficient search for the optimal parameters of the logistic transformation\footnote{In machine learning parlance, the new score can broadly be categorized as a generalized additive model \cite{hastie_2009}.}. In particular, for patient $i$ and feature $j$, the nonlinear 
transformation $z_{ij}$ of the raw variable $x_{ij}$ is
\begin{equation} 
\label{eq:nl}
z_{ij} = \begin{cases} \frac{1}{1+\exp(-a_j(x_{ij}-t_{ij}))} & \mbox{if } x_{ij} \mbox{ is a maximum} \\ 
1-\frac{1}{1+\exp(-a_j(x_{ij}-t_{ij}))} & \mbox{if } x_{ij} \mbox{ is a minimum} \\
0 & \mbox{if } x_{ij} \mbox{ is missing} \\ \end{cases}
\end{equation} 
where $a_j\ge$0 is the slope of the nonlinear transformation and $t_{ij}$ is the inflection point of 
the logistic function (i.e., a ``soft threshold" counterpart to the hard thresholds used by the existing 
stepwise scoring schemes). It should be noted that if clinical 
knowledge suggests the use of age-dependent thresholds, the $t_{ij}$'s are not different for every subject 
$i$ but rather have the same value for all patients within a specific age group and a given feature 
$j$. In the case where this age-dependence is not evidenced, the soft thresholds $t_{ij}$ in the novel algorithm 
are shared across the entire patient population and thus the soft threshold for feature $j$ can be written as 
$t_j$.

The optimal weights and parameters for the nonlinear transformations are determined 
by minimizing the negative log-likelihood of the logistic regression model, 
\begin{equation}
\label{eq:obj}
\min \sum_{i=1}^{n} \log(1+\exp(-y_i(\mathbf{w}^T{\mathbf{z}_i}))).
\end{equation}
To preserve and exploit clinical knowledge previously used in the creation of other scores, a lognormal prior is 
imposed in the optimization for $\mathbf{w}$; this also ensures all features will be associated with a positive 
weight. In particular, for $\mathbf{w}\in \mathbb{R}^d$ we set 
$$ \mathbf{P}(\mathbf{w})=\frac{\exp\left( -\frac{1}{2}(\log \mathbf{w}-\mu)^T\Sigma^{-1}(\log \mathbf{w}-\mu) \right)}{(2\pi)^{d/2}\vert\Sigma\vert^{0.5}\prod\limits_{j=1}^d w_j}.$$
For a lognormal prior with mean $\mu$ and covariance $\Sigma=\frac{1}{2\lambda}I$, the optimization over $\mathbf{w}$ then becomes
\begin{align}
\arg\min\limits_\mathbf{w} \sum_{i=1}^{n} \log(1+\exp(-y_i(\mathbf{w}^T{\mathbf{z}_i}))) +\sum\limits_{j=1}^d\log{w_j}+\lambda\Vert\log\mathbf{w}-\mu\Vert_2^2. \label{eq:wlognorm}
\end{align}

The joint optimization (\ref{eq:wlognorm}) over $\mathbf{w}$, 
$\mathbi{a}=[a_1 \; a_2 \; \ldots \;a_{M_a}]$, and/or  $\mathbf{t}=[t_{1} \; t_2 \; \ldots \; t_{M_t}]$ is 
carried out by cyclic block coordinate descent with backtracking line search \cite{boyd_2009}. 
Optimization over $\mathbi{a}$ and $\mathbf{t}$ includes an additional step of projections onto 
the constraint set. The blocks for the coordinate descent consist of features derived from the same 
raw variable. For example, one block contains two features derived from the maximum heart rate, 
which correspond to two steps with different weights in a simple thresholding-based additive score.
The algorithm is formalized as Algorithm \ref{alg:opt_a} given below. Note that the objective function 
of the optimization (\ref{eq:wlognorm}) is not convex. Nevertheless, even if we use an iterative 
optimization method that merely ensures the objective function is decreased at each step, in our 
empirical studies the resulting local minimum leads to an improvement over the existing 
risk score. In fact, our computational studies show that the proposed scheme is robust with respect 
to the initial point of the search -- for instance, in the application to pediatric population, starting the 
iterative optimization procedure with a vector $\mathbf{w}$ comprising PRISM III weights and starting 
with a vector of uniform weights $\mathbf{w}=([1,\cdots,1]$ results in almost identical prediction 
accuracy on the considered dataset.

\begin{algorithm}
\caption{Optimization over the slopes $\mathbi{a}$}
\label{alg:opt_a}
\begin{algorithmic}
	\STATE $ a_j^{(0)}\leftarrow 0.01,\ j=1,2,...,M_a$
	\STATE $\mathbf{w} \leftarrow \mathbf{w}_{\mathrm{PRISM}} $
	\STATE $\mathbf{t} \leftarrow \mathbf{t}_{\mathrm{PRISM}} $
	\STATE $k \leftarrow 1 $
	\REPEAT 
	\STATE $\mathbi{a} \leftarrow \mathbi{a}^{(k-1)}$
	\FORALL { $g \in G$}
		%\STATE $f \leftarrow f(\mathbi{a}, \mathbf{t}, \mathbf{w}) $
		\STATE $\triangle a_j \leftarrow -(\nabla_af(\mathbi{a}, \mathbf{t}, \mathbf{w}))_j $ if $j \in g$ 
		\STATE $\triangle a_j \leftarrow 0 $ if $j \not\in g$ 
		\STATE $ h \leftarrow 1 $
		\WHILE{$ f(\mathbi{a}+h\triangle\mathbi{a}, \mathbf{t}, \mathbf{w})>f(\mathbi{a}, \mathbf{t}, \mathbf{w})-\alpha h ||\triangle \mathbi{a} ||^2$}
			\STATE $h \leftarrow \beta h$
		\ENDWHILE
		\STATE $\mathbi{a} \leftarrow \mathrm{Proj}_A(\mathbi{a}+h\triangle\mathbi{a}) $
	\ENDFOR
	\STATE $ \mathbi{a}^{(k)} \leftarrow \mathbi{a} $
	\UNTIL stopping criterion is met
\end{algorithmic}
\end{algorithm}

In Algorithm \ref{alg:opt_a}, $G$ is the set of feature groups, $A$ denotes the projection set for 
$\mathbi{a}$, and $\alpha$ and $\beta$ are the backtracking 
line search parameters. The feature groups $g \in G$ are defined as the groups of nonlinear 
features related to the same raw variables (e.g., in the application to pediatric population the 
feature group for maximum pH has two elements 
corresponding to PRISM III thresholds of $7.55$ and $7.48$), and make up the blocks for the 
coordinate descent. 

The projection sets are defined so that the clinical knowledge used for the existing score is preserved. 
In Algorithm \ref{alg:opt_a}, $A$ ensures the nonlinear transformations follow the same direction 
as the steps in the existing score ($A=\{ \mathbi{a} : a_j \ge 0 $ for $j=1,\dots,M_a\} $). The projection 
set $T$ for optimization over the soft thresholds of the nonlinear transformations in (\ref{eq:nl}), 
$\mathbf{t}$, preserves the order of soft thresholds for a raw variable 
with multiple nonlinear transformations. For example, if the feature $i$ in SOFA corresponds to 
the minimum platelet count between $100$ and $150$ $\times 10^3/\text{mm}^3$ ($t_i^{(0)}=150$) and 
feature $j$ corresponds to the minimum platelet count between $50$ and $100$ $\times 10^3/\text{mm}^3$ ($t_j^{(0)}=100$), 
then the projection ensures that $t_i \geq t_j$ at all steps of the optimization procedure. It should be noted that this projection set may lead to a collapse of thresholds if the optimal solution is 
found at $t_j=t_i$. For example, the minimum platelet count, which has 4 levels in SOFA, may 
end up having only two different inflection points in the novel score.

The dimensions of the three optimization parameters are not equal due to the existence of binary 
features (e.g., those associated with pupillary reaction in PRISM III and drug 
administration in SOFA) that do not have nonlinear transformations ($M_a<M_w$) and the 
age-dependence of some of the thresholds ($M_w<M_t$). Optimization over the slopes 
$\mathbi{a}$ and soft thresholds $\mathbf{t}$ are implemented without inclusion of a prior in 
the objective. 

\subsection{Quasiconvexity of the logistic transformation}

As stated earlier in this section, the objective function in (\ref{eq:wlognorm}) is not convex. 
However, we will here show that each block in the block-coordinate descent procedure is 
both quasiconvex and quasiconcave in the slope parameter $\mathbi{a}$, and is thus quasilinear. 
Since the logistic function is asymptotically flat, the objective in (\ref{eq:wlognorm}) is not 
strictly quasiconvex. However, since the objectives in the steps of block-coordinate descent 
procedure are quasilinear, if the initial values of $\mathbi{a}$ are such that the gradient is nonzero in 
every coordinate, the block coordinate descent will reach a global optimum. 

For differentiable $f$ and domain $\mathcal{D}$, $f$ is quasiconvex if and only if 
$\mathcal{D}$ is convex and for all $x,y \in \mathcal{D}$ holds that 
$f(y) \leq f(x) \Rightarrow \nabla{f(x)}^T(y-x) \leq 0$ \cite{boyd_2009}. Similarly, $f$ is 
quasiconcave if and only if $\mathcal{D}$ is convex and for all $x,y \in \mathcal{D}$
it holds that $f(y) \geq f(x) \Rightarrow \nabla{f(x)}^T(y-x) \geq 0$.

The objective of the optimization is
\begin{equation}
\label{eq:obj2}
\min f=\sum_{i=1}^{n} \log(1+\exp(-y_i(\mathbf{w}^T{\mathbf{z}_i}))),
\end{equation}
where 
\begin{align}
\mathbf{w}^T\mathbf{z}_i= \sum_{j \in U}\frac{w_j}{1+\exp(-a_j(x_{ij}-t_{ij}))} + \sum_{k \in D}w_k \left( 1-\frac{1}{1+\exp(-a_k(x_{ik}-t_{ik}))} \right) + \sum_{p \in P} w_p \delta(p_i=1),
\end{align}
where $U$ denotes the set of indices of features with maximum values whose contribution 
to the score is in the form of an up-step (i.e., the risk is higher when their values are above 
a threshold), $D$ is the set of indices of features with minimum values (down-steps), $P$ 
is the set of indices 
of the pupillary reflex features taking binary values $\{0,1\}$, and $\delta(p_i=1)$ is an indicator function for the pupillary reflex features. 
The pupillary reflex features indicate whether one or both pupils are $>3$mm and 
fixed.

Note that if we want to find when is $f(a_j')\leq f(a_j'')$ for a given $a_j'$ and $a_j''$, 
it is sufficient to find the conditions on $a_j', a_j'', y_i, x_{ij}, t_{ij}$ such that it holds
that $f_i(a_j')\leq f_i(a_j'')$ for all $i \in {1, \ldots n}$, where $f=\sum_{i=1}^n f_i$. 
For any $j \in U$, condition $f(a_j') \leq f(a_j'')$ is satisfied on the domain where
$$  \frac{-y_i w_j}{1+\exp(-a_j'(x_{ij}-t_{ij}))} \leq   \frac{-y_i w_j}{1+\exp(-a_j''(x_{ij}-t_{ij}))} .$$
This inequality will hold for any of the parameter combinations marked by an X in 
Table \ref{table:aquasi1}.
\begin{table}[!h]
%% increase table row spacing, adjust to taste
\renewcommand{\arraystretch}{1.3}
\caption{Satisfy $f_i(a_j')\leq f_i(a_j'')$ for $j \in U$}
\label{table:aquasi1}
\begin{center}
\begin{tabular}{ |c|c|c|c|c| }
\cline{2-5}
\multicolumn{1}{c|}{} & \multicolumn{2}{c|}{$a_j' \leq a_j''$} & \multicolumn{2}{c|}{$a_j' \geq a_j''$} \\
\cline{2-5}
\multicolumn{1}{c|}{} & $y_i=1$ & $y_i=-1$ & $y_i=1$ & $y_i=-1$ \\
\cline{1-5}
$x_{ij}-t_{ij} \geq 0$ &   & X & X &   \\
\hline
$x_{ij}-t_{ij} \leq 0$  & X &  &  &  X  \\
\hline
\end{tabular}
\end{center}
\end{table}
The blank spaces in Table \ref{table:aquasi1} satisfy $f_i(a_j')\geq f_i(a_j'')$ for $j \in U$, the fact
which we next use to show quasiconcavity. To show quasilinearity, we also need to examine the 
gradient of $f$ at $a_j''$ which, for $j \in U$, is given by
\begin{align}
\label{eq:gradf}
\nabla_{a_j}f_i(a_j'')= -\frac{\exp(-y_i(\mathbf{w}^T{\mathbf{z}_i}))}{1+\exp(-y_i(\mathbf{w}^T{\mathbf{z}_i}))} \times  \frac{y_i w_j (x_{ij}-t_{ij})\exp(-a_j''(x_{ij}-t_{ij}))}{\left(1+\exp(-a_j''(x_{ij}-t_{ij}))\right)^2}.
\end{align}
The sign of the gradient in (\ref{eq:gradf}) is determined by $-y_iw_j(x_{ij}-t_{ij})$. Simple arithmetic 
shows that when $f_i(a_j')\leq f_i(a_j'')$ (corresponding to the entries marked by X in 
Table \ref{table:aquasi1}), $\nabla_{a_j}f_i(a_j'')(a_j'-a_j'') \leq 0$, and hence the condition for
quasiconvexity is satisfied. Similarly, when $f_i(a_j')\geq f_i(a_j'')$ (corresponding to blanks in 
Table \ref{table:aquasi1}), $\nabla_{a_j}f_i(a_j'')(a_j'-a_j'') \geq 0$, which implies quasiconcavity.

The same procedure can be followed to show quasilinearity when $j \in D$, with appropriate sign 
changes. Note that the blocks of features where all of the features in the block belong to $U$ or 
all of the features in the block belong to $D$ will also satisfy the 
quasilinearity condition. Since the coordinate blocks in Algorithm \ref{alg:opt_a} correspond to 
nonlinear transformations of the same physiological variable, quasilinearity holds. The initial value 
of $a_j$ in Algorithm \ref{alg:opt_a} is set to $0.01$ for all $j$. This corresponds to a small slope 
in the nonlinear transformations of the variables, and will only result in $\nabla_{a_j}f_i(a_j)=0$ 
for subjects $i$ with $x_{ij}=t_{ij}$. Therefore, the initial slope of the cumulative function in 
(\ref{eq:obj2}) will be nonzero and the coordinate descent algorithm will not begin at a stationary 
point.

Since we constrain the slopes $\mathbi{a}$ and weights $\mathbf{w}$ to be nonnegative, quasilinearity 
of the objective function in soft thresholds $\mathbf{t}$ can be shown in fewer steps than that for
$\mathbi{a}$. We will only show quasiconvexity in $t_{ij}$ when $j\in U$, but quasiconvexity when 
$j\in D$ and quasiconcavity can be easily shown in a similar manner. To demonstrate quasiconvexity, 
it is sufficient to show that 
\[
f_i(t_{ij}') \leq f_i(t_{ij}'') \Rightarrow \nabla_{t_{ij}}f_i(t_{ij}'')(t_{ij}'-t_{ij}'') \leq 0.
\]
Note that for $j\in U$, $f_i(t_{ij}') \leq f_i(t_{ij}'')$ if $y_i=-1\text{ and }t_{ij}' \geq t_{ij}''$, or if 
$y_i=1\text{ and }t_{ij}' \leq t_{ij}''$. 
By examining the gradient of $f_i$, it is clear that since $w_j,a_j\geq 0$, $\nabla_{t_{ij}}f_i(t_{ij}'') \leq 0$ 
only when $y_i=-1$,

\begin{align}
\label{eq:gradft}
\nabla_{t_{ij}}f_i(t_{ij}'')=\frac{\exp(-y_i(\mathbf{w}^T{\mathbf{z}_i}))}{1+\exp(-y_i(\mathbf{w}^T{\mathbf{z}_i}))} \times  \frac{y_i w_j a_j\exp(-a_j'(x_{ij}-t_{ij}''))}{\left(1+\exp(-a_j(x_{ij}-t_{ij}''))\right)^2}.
\end{align}

Note that the required conditions for $f_i(t_{ij}') \leq f_i(t_{ij}'')$ also guarantee that
$\nabla_{t_{ij}}f_i(t_{ij}'')(t_{ij}'-t_{ij}'') \leq 0$. However, unlike with the optimization over slopes 
$\mathbi{a}$, we cannot guarantee that the initial $\mathbf{t}$ will be at a point with nonzero 
slope. Nevertheless, we have empirically observed that although $\nabla_{t_{ij}}f_i=0$ for some 
$(i,j)$, $\sum_i\nabla_{t_{ij}}f_i \neq 0$ in the first iteration of the alternating optimization of 
$\mathbi{a}$ and $\mathbf{t}$, indicating the optimization does not start at a stationary point for any 
of the nonlinear features. As with the weights $\mathbf{w}$, we initialize $\mathbf{t}^{(0)}$ using
the hard thresholds of the existing score. Despite the approximations made in order to simplify calculation of the existing score, one expects its thresholds to be close to the optimal values due to the reliance
on domain expert knowledge and extensive testing. Therefore, the proposed initialization will likely avoid 
start of the optimization procedure in the flat part of the quasi convex curve, implying the global
optimality of the block coordinate descent.

\subsection{Algorithm testing} \label{sec:testing}
We validate the algorithm on a pediatric brain injury dataset and an adult ICU dataset. 
Note that the proposed method allows a straightforward refinement and optimization of the novel 
risk prediction score for a specific subpopulation and/or location of the hospital. The proposed 
prediction scheme is tested using leave-one-out cross-validation for the pediatric 
dataset (n=217) and 10-fold cross-validation for the adult dataset (n=3711) and compared
with existing methods in terms of three discrimination criteria: (i) area under the receiver 
operating characteristic 
(ROC) curve (AUC), (ii) the Youden index ($J$), which aims to maximize the overall correct 
classification rate \cite{hilden_1996,perkins_2006}: 
$J=\mathrm{Sensitivity }(Se)+ \mathrm{Specificity}(Sp) -1$, and (iii) the point on the ROC curve 
that maximizes the minimum of the positive predictivity ($+P$, precision) and sensitivity (recall). 
The last criterion takes into account class imbalance by balancing the percentage of true positives 
that are correctly predicted with the percentage of predicted positives that are correct. Thus, 
correctly predicted true negatives, the majority class, does not affect the performance metric. 
It should be noted that in evaluating algorithms by the third criterion (PrecRec), a false positive 
has the same effect as a false negative. These criteria only assess the 
discriminatory capabilities of the algorithms. In prognostic models, particularly those that lead to 
clinical decision-making, the preciseness of the predicted probability, or calibration, should also 
be examined \cite{clinical, medlock_2011}.  We assess the model calibration in terms of the 
Brier score (BS) \cite{brier_1950}, calculated as 
\begin{equation}
BS = \frac{1}{N} \sum\limits_{k=1}^N(\pi_k-c_k),
\end{equation}
where $N$ denotes the number of subjects, $c_k\in\{0,1\}$ is the binary class for subject $k$, and $\pi_k$ is the probability of mortality for subject $k$. The probabilities were calculated by means of Platt scaling \cite{platt_1999} during the cross-validation, such that the algorithm outputs were entered in a logistic regression to find the probability of mortality. The Brier score ranges from 0 to 1, where lower values indicate a better calibrated score.

\section{Results and Discussion} \label{sec:results}
In this section, we analyze pediatric and adult datasets to demonstrate the performance 
of the novel scoring scheme that preserves expert knowledge from PRISM III and SOFA, 
respectively. Therefore, the variables used by the novel score and the initial parameters for 
the optimization (feature weights and thresholds) are those from the aforementioned existing 
scores.

\subsection{Implementation and results for the novel pediatric population scores}\label{sec:resultsPedi}
Here we detail the implementation of the novel algorithm and present 
the discrimination and calibration results from the leave-one-out cross-validation tests on a 
population of 217 children with brain trauma and other brain malady or injury. We compare
these results with those achieved by the classifiers that use raw non-transformed data as 
features. Since the latter classifiers do not allow for missing data (as in the case of PRISM 
III and our novel algorithm), different imputation strategies are examined.

\subsubsection{Pediatric patient population}
Data were retrospectively abstracted for 217 children (11.06\% mortality rate) admitted to the 
Dell Children's Medical Center PICU. We included admissions to the PICU between 
August 2007 and April 2012, age range of 0-14 years, with an ICD-9 code reflecting brain injury, 
and a PICU stay of at least 24 hours. The ICD-9 codes indicate patients with brain trauma 
(excluding simple concussion) as well as other brain malady or injury such as cerebral palsy, 
drowning, epilepsy and  asphyxiation. This group of patients was selected in order to emphasize 
the ease of optimizing the novel score for a specific high mortality population. This particular high-mortality population is of interest since trauma is the leading cause of death in children in the United States \cite{murphy_2012}, with traumatic brain injury being a major contributor \cite{faul_2010}.
Minimum and maximum variable values from the first 
$12$ hours of the PICU stay are used in the calculation of PRISM III and our new score. The 
use of the data has been approved by the University of Texas at Austin Institutional Review 
Board and the Seton Clinical Research Steering Committee. 

\subsubsection{Implementation of the novel prediction scheme in pediatric population} \label{sec:implPedi}
Our novel score incorporates all of the variables and ranges previously used by PRISM III. 
Specifically, the 17 variables that our algorithm uses include: systolic blood pressure, 
heart rate, body temperature, pupillary reflexes, Glasgow Coma Scale, total CO\textsubscript{2}, 
pH, PaO\textsubscript{2}, PCO\textsubscript{2}, glucose, potassium, creatinine, blood urea 
nitrogen, white blood cell count, platelet count, PT, and PTT. Detailed information about the ranges 
and cutoff points for PRISM III can be found in \cite{pollack_1996}. The 
features transformed using OR statements in PRISM III (e.g., by adding 6 points to the score if 
$\mathrm{pH}<7.0$ OR $\mathrm{total \ CO}_2<5 $) were treated as additive features in order 
to seamlessly include them in the optimization procedure. The backtracking line search 
parameters were chosen empirically based on the discrimination and the speed of convergence 
of the algorithm. The 
results shown are for $\alpha=0.2$ and $\beta=0.5 $. In the case of optimization over 
$\mathbf{w}$, the imposed prior has $\mu=0, \lambda=0.25$.

\subsubsection{Performance comparison of the novel scores and existing scores}
\begin{table}
%% increase table row spacing, adjust to taste
\renewcommand{\arraystretch}{1.3}
% if using array.sty, it might be a good idea to tweak the value of
% \extrarowheight as needed to properly center the text within the cells
\caption{Mortality and Missing Data by Age Group}
\label{patientChar}
\centering
%% Some packages, such as MDW tools, offer better commands for making tables
%% than the plain LaTeX2e tabular which is used here.
\begin{centering}
\begin{tabular}{|c|c|c|m{2.5cm}|}
\hline
Age Group & No. Subjects & No. Deaths & Missing Variables\par (mean $\pm$ SD)\\
\hline
Neonate (0 mo., 1 mo.) & 2 & 0 & 9 $\pm$ 11.31\\
\hline
Infant [1 mo., 12 mo.) & 39 & 5 & 6.95 $\pm$ 7.13\\
\hline
Child [12 mo., 144 mo.] & 143 & 15 & 7.02 $\pm$ 7.22\\
\hline
Adolescent (144 mo., 180 mo.) & 33 & 4 & 6.03 $\pm$ 7.14\\
\hline
\end{tabular}
\end{centering}
\end{table}

%NEED TO UPDATE DATA FOR THIS GRAPH
\begin{figure}[!h]
\begin{centering}
%centering
\includegraphics[width=3.5in]{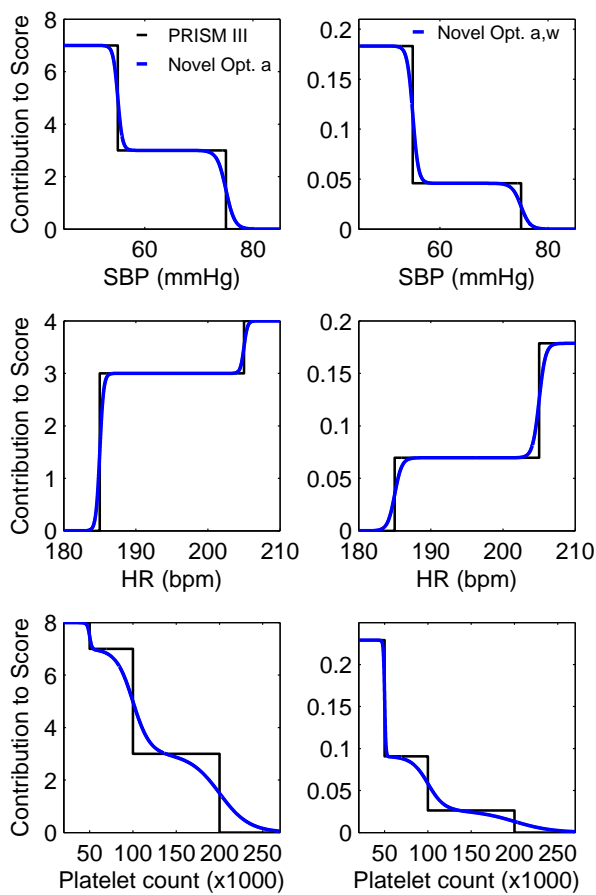}
 %where an .eps filename suffix will be assumed under latex, 
% and a .pdf suffix will be assumed for pdflatex; or what has been declared
% via \DeclareGraphicsExtensions.
\caption{Contribution of systolic blood pressure (SBP), heart rate (HR), and platelet count to risk score for PRISM III (black, left) and novel score optimizing $\mathbi{a}$ (blue, left) and optimizing $\mathbi{a}, \mathbf{w}$ (blue,right). The black lines on the right panel serve as indicators of a sharp transition between steps.}
\label{fig_newSlopes}
\end{centering}
\end{figure}

Figure \ref{fig_newSlopes} illustrates the difference in computing the contribution of a feature to the 
prediction score between the $12$-hour PRISM III and the novel scheme. Plots on the left show the
logistic transformations of the features after performing optimization over the nonlinear transformation 
slopes $\mathbi{a}$, while the plots on the right show the logistic transformations after performing
optimization over both the slopes $\mathbi{a}$ and weights $\mathbf{w}$. From the plots, we see
that the optimization over $\mathbi{a}$ results in mortality risk increasing over a range of 3 to 4 
beats per minute for each step in the maximum heart rate and the risk from minimum systolic blood pressure 
increasing over a range of 7 mmHg around the 75 mmHg threshold and increasing over a range of 4 mmHg around the 55 mmHg threshold. The risk from the minimum platelet count loses much of the 
stepwise scoring structure used by PRISM III, and instead increases monotonically for minimum platelet 
counts between 250,000 and 50,000. This illustrates how our novel score can capture risk 
that increases continuously throughout a certain range while maintaining sharp thresholds 
when those are optimal. Optimizing over both $\mathbi{a}$ and $\mathbf{w}$ provides further 
insight into the contribution of variables to the risk of mortality for a given population. For 
example, the novel score shows that, for this dataset, the second step in the nonlinear 
transformation of the systolic blood pressure and heart rate should be weighted similarly, 
if not more heavily, than the lower step. The lower panels of Figure \ref{fig_newSlopes} also 
indicate that while PRISM III has the mortality risk increasing slightly more when the minimum 
platelet count falls below 50,000, our inferred scoring function indicates the risk contribution 
from this variable doubles at approximately 50,000. 

\begin{table}
%% increase table row spacing, adjust to taste
\renewcommand{\arraystretch}{1.3}
%% if using array.sty, it might be a good idea to tweak the value of
%% \extrarowheight as needed to properly center the text within the cells
\caption{Risk Score Accuracy: Pediatric Population}
\label{ROCtable}
\centering
%% Some packages, such as MDW tools, offer better commands for making tables
%% than the plain LaTeX2e tabular which is used here.
\begin{center}
\begin{tabular}{|c|c|c|c|c|}
\hline
Score & AUC & $J$ & PrecRec & BS \\
\hline
PRISM III &  0.8735 &  0.5840 & 0.5172 &  0.0686\\
\hline
Novel Score optimized over $\mathbi{a}$ & 0.8897  & 0.6215  & 0.5000 &  0.0680\\
\hline
Novel Score optimized over $\mathbf{w}$ &  0.8841& 0.6369  & 0.5600 & 0.0715\\
\hline
Novel Score optimized over $\mathbi{a}$ and $\mathbf{t}$ & 0.8358 & 0.6153   & 0.5833 & 0.0708\\
\hline
Novel Score optimized over $\mathbi{a}$ and $\mathbf{w}$ &  0.8927 &  0.6682 & 0.5833 & 0.0679\\
\hline
PIM 2 & 0.8331 & 0.6729 & 0.5417 &  0.0766\\
\hline
\end{tabular}
\end{center}
\end{table}

\begin{figure}[!h]
\begin{centering}
\includegraphics[width=5.5in]{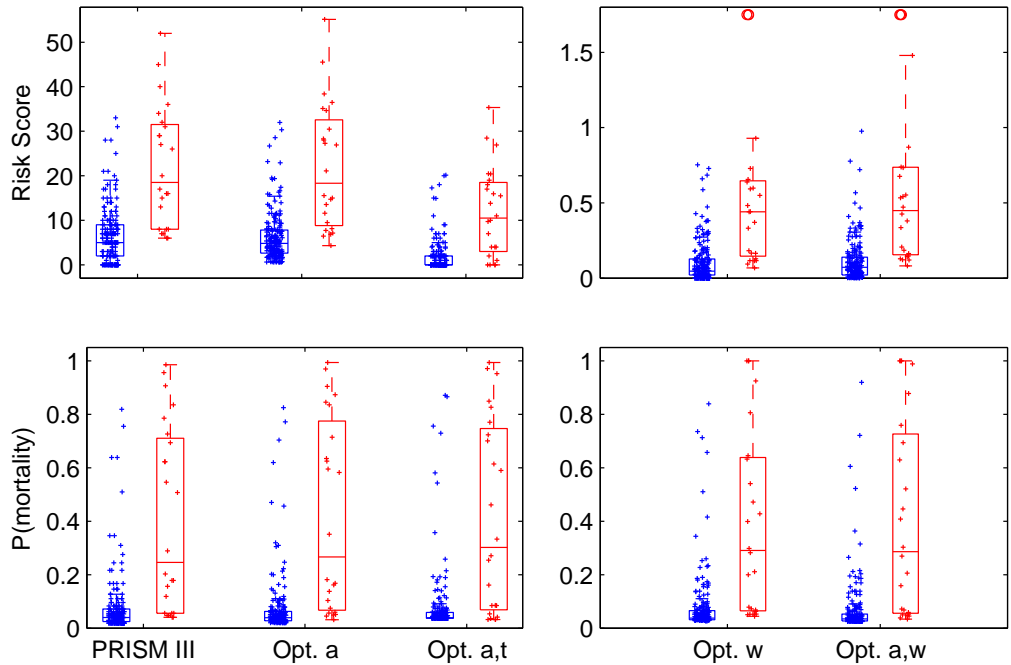}
 %where an .eps filename suffix will be assumed under latex, 
% and a .pdf suffix will be assumed for pdflatex; or what has been declared
% via \DeclareGraphicsExtensions.
\caption{Risk scores and the probability of mortality for survivors (blue) and 
nonsurvivors (red). The center mark on the box 
indicates the median while the edges of the box mark the 25th and 75th percentile. Individual scores 
are plotted as +. In the top right panel, the red circles indicate subjects with scores higher than 2.}
\label{fig_riskGroup}
\end{centering}
\end{figure}

The results of the ROC analysis on the cross-validated scores are shown in 
Table \ref{ROCtable}. Our novel score optimized over the slopes $\mathbi{a}$ of the 
nonlinear transformations results in a more accurate classifier than PRISM III, in terms of 
the AUC and $J$, but not in terms of the precision-recall balance (PrecRec). Optimization 
of the feature weights $\mathbf{w}$ results in a score that performs better than PRISM III
in all three evaluation criteria. Alternating optimization over the slopes $\mathbi{a}$ 
and weights $\mathbf{w}$ results in a further improvement of AUC, $J$ and precision-recall 
balance over individual parameter optimization, as well as the best calibration 
values (Brier score), and thus provides a significant advancement
over PRISM III. The inclusion of soft thresholds $\mathbf{t}$ along with slopes 
$\mathbi{a}$ in the alternating optimization results in some of the soft thresholds falling 
outside the physiological range of the raw variables -- in particular, logistic
transformations of these variables result in zero contribution to the risk score for all 
patients. This optimization results in better discrimination than PRISM III in the upper range 
of scores (higher $J$ and precision-recall balance) and poorer classification than PRISM 
in the lower score range (lower AUC). However, we expect that these results would improve 
with a larger patient population given the age-dependency of some of thresholds and 
the mortality distribution across age groups (Table \ref{patientChar}). The ROC results for 
PIM2 are also included in Table \ref{ROCtable} to compare the novel score to another 
widely used pediatric risk score. The novel score optimized over any of the parameters 
outperforms PIM2 in terms of AUC, and optimization over 2 parameters also yields a higher 
precision-recall balance than PIM2. Though PIM2 results in a slightly higher $J$ than 
the novel score optimized over $\mathbi{a},\mathbf{w}$, the large gain in AUC and higher 
precision-recall balance make the novel score the preferred choice for predicting risk of 
in-hospital mortality in the studied population.

The boxplots in Figure \ref{fig_riskGroup} illustrate how the proposed prediction scheme
compares with PRISM III. The probability of mortality (right column 
in Figure \ref{fig_riskGroup}) is also included in order to compare the algorithms on the 
same scale. The probability of mortality is calculated by means of Platt scaling as stated 
in Section \ref{sec:testing}. Despite additional features used by our scheme (due to splitting 
of the OR statements into components), the novel scores have similar average values 
to those of PRISM III. Given the class imbalance, this is likely the result of lower slopes in 
the nonlinear transformations which decrease the feature contributions of survivors 
having measurements near the soft thresholds. The movement of soft thresholds outside the 
physiological range following inclusion of $\mathbf{t}$ as an optimization parameter 
results in lower average score values compared to PRISM III. Finally, reduction in mean 
scores when the weights are included as optimization variables is expected due to the 
prior on the weight distribution. It should be noted that the optimizations which 
result in overall lower risk scores (i.e., optimizations including $\mathbf{t}$ or $\mathbf{w}$) 
yield lower risk scores for both the survivor and nonsurvivor groups; the relative difference 
between the groups remains and the mean probability of mortality is not significantly different 
from that of PRISM III or the score optimized only over $\mathbi{a}$.

\begin{figure}[!h]
\begin{centering}
\includegraphics[width=4.5in]{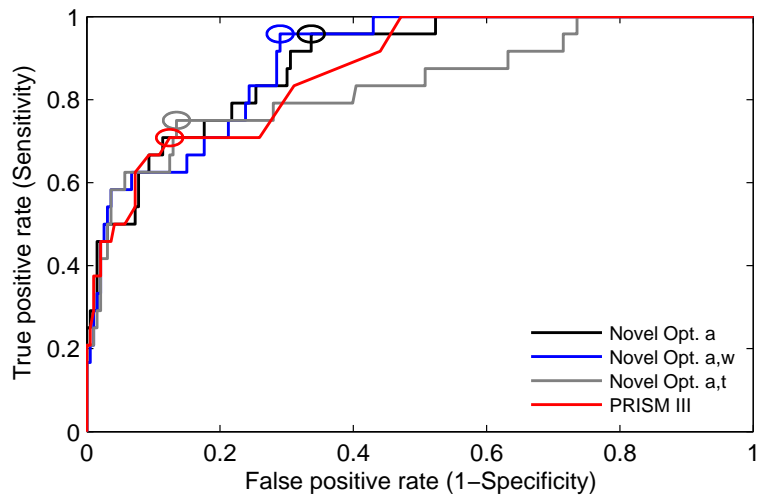}
 %where an .eps filename suffix will be assumed under latex, 
% and a .pdf suffix will be assumed for pdflatex; or what has been declared
% via \DeclareGraphicsExtensions.
\caption{ROC curve for PRISM III (red) and novel scores optimized over the nonlinear 
transformation slopes $\mathbi{a}$ (black), and alternating minimization over the slopes 
$\mathbi{a}$ and weights 
$\mathbf{w}$ (blue) and the slopes $\mathbi{a}$ and soft thresholds
$\mathbf{t}$ (gray) of the nonlinear transformations. The circled points correspond to those 
that maximize Youden's index.}
\label{fig_ROCa}
\end{centering}
\end{figure}

The difference in AUC between PRISM III and the novel scores (Figure \ref{fig_ROCa}) 
is primarily caused by low PRISM III scores for some nonsurvivors. Due to the finite set 
of PRISM III values, patients are more likely to share the same score which causes
decrease in both the true positive rate and the false positive rate as the cutoff value is 
lowered in the ROC analysis. The largest differences in the curves can be traced to 
patients with PRISM III scores between 7 and 13. While 45 out of 193 survivors have 
PRISM III scores in this range, so do 7 out of 24 nonsurvivors. We further 
examined the effect of softening the thresholds on patient outcome prediction for the 
subjects with low PRISM values by selecting the thresholds corresponding to the highest 
specificity with a sensitivity of at least 90. Such a sensitivity restriction promotes accurate
identification of the in-hospital mortality class. For the selected cutoff scores, the novel 
score (optimized over the slopes $\mathbi{a}$) correctly identifies three subjects as high 
risk of in-hospital mortality that are incorrectly classified by PRISM III. One of these 
subjects is correctly identified by the novel score due to a nonzero contribution to the 
score from a systolic blood pressure with a value precisely at the PRISM III hard threshold. 
The other two subjects are correctly identified by the novel score due to nonzero 
contributions from the Glasgow Coma Scale feature ($\text{GCS}\in\{8,9\}$), which in the 
optimized score has an approximately linear relationship to risk whereas PRISM III is 
affected only if $\text{GCS}<8$.

\subsubsection{Comparison of the novel and existing scores for pediatric age groups}
\begin{figure}[!h]
\begin{centering}
\includegraphics[width=5.5in]{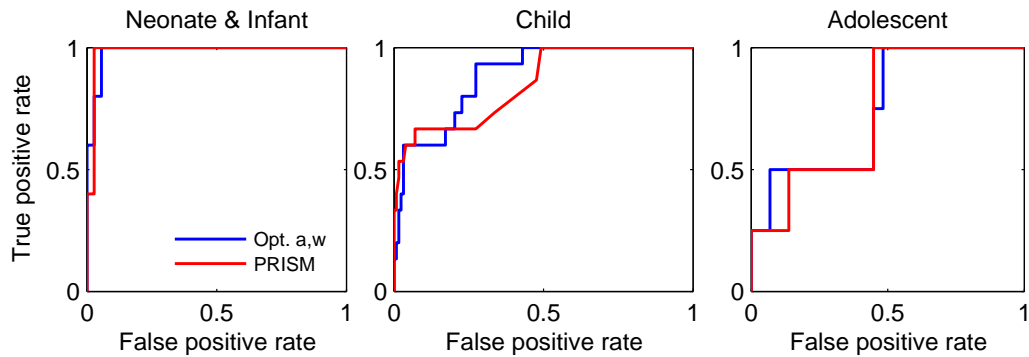}
 %where an .eps filename suffix will be assumed under latex, 
% and a .pdf suffix will be assumed for pdflatex; or what has been declared
% via \DeclareGraphicsExtensions.
\caption{The ROC curve by age group for PRISM III (red) and the novel score 
optimized over the slopes $\mathbi{a}$ and weights $\mathbf{w}$ of the proposed nonlinear 
transformation (blue).}
\label{fig_ROCage}
\end{centering}
\end{figure}

Analysis of the results for different age groups suggests that the novel algorithm provides
the largest gains in discrimination for the children in the age group (12-144 months). The ROC 
curves for different age groups are shown in Figure \ref{fig_ROCage}, where the neonate 
and infant groups are combined due to the low number of subjects in the latter group. 
 Note that the presented results display performance of the novel algorithm 
optimized over the entire population rather than those of three separate optimizations using 
only subjects from the age groups of interest. For the neonate and infant groups 
(n=41, 5 deaths), both PRISM III and the novel score optimized over the slopes 
$\mathbi{a}$ and weights $\mathbf{w}$ of the nonlinear transformation achieve the AUC of 
0.9833. In the child age group (n=143, 15 deaths), the novel score (AUC=0.8844) 
was significantly more discriminative than PRISM III (AUC=0.8516). Both scores 
exhibited less accurate performance in the adolescent group (n=33, 4 deaths), with 
the novel score (AUC=0.7500) slightly outperforming PRISM III (0.7414). 
The high AUC in the neonate and infant group can be attributed to the 
characterization of the subjects in the nonsurvivor group, where most of them were near 
drowning victims and unresponsive upon ICU admission. The worse discrimination in the 
adolescent group is likely due to the low number of subjects and deaths, where two of the 
subjects in the nonsurvivor group had low scores with both the novel algorithm and PRISM III. 
 It should also be noted that the distribution of brain injuries in the adolescent group 
is different from the other two groups. For example, this age group has a higher incidence of 
the asphyxiation/strangulation ICD-9 code. Given that the method presented in this paper can 
be used to design optimal scores for different populations, were more data available, it would 
have been interesting to design a mortality prediction score specifically for the adolescent 
population.

\subsubsection{Consideration of alternate feature transformation}
To illustrate the improvement in discrimination enabled by using nonlinear transformations of 
the features, we also performed a logistic regression with ridge, lasso \cite{tibshiriani_1994}, 
and elastic net \cite{zou_2005} penalties on the age and raw variables of the datasets. Additionally, to demonstrate the significance of the specific logistic feature 
transformations used by the novel score, we implemented and tested logistic regression 
models with restricted cubic spline transformation of the raw physiological variables. Since 
missing data is prevalent and complete data sets are required for logistic regression, missing 
values of the variables are imputed by $k$ nearest-neighbors ($k$NN) \cite{hastie_1999}, the 
probabilistic principal components analysis (PPCA) method \cite{tipping_2002}, mean values 
and normal values \cite{behrman_2004}. In the $k$NN imputation, Euclidean distances 
normalized by the number of common features between patients are calculated and the 
missing values are imputed as the average value of the $k$-nearest neighbors that observe 
the variable. PPCA aims to reduce the dimensionality of the data by associating a Gaussian 
latent variable model with the observed data and imputing missing values by an iterative 
expectation-maximization procedure. In our tests, data were imputed with $k=5$ and $4$ 
principal components. The best AUC values were achieved using the elastic net 
for the linear feature set and elastic net or ridge penalty for the nonlinear feature set. The 
discrimination and calibration measures for the logistic regression with linear (raw) 
and nonlinear (cubic spline transformation) feature sets are compared to the PRISM III 
values calculated with the imputed dataset in Table \ref{table:rawResults}. 

\begin{table}
\renewcommand{\arraystretch}{1.3}
\caption{Risk Score Accuracy: Logistic with Raw Variables vs. PRISM with Imputed Data}
\label{table:rawResults}
\begin{center}
\begin{tabular}{ |c|c|c|c|c|c| }
\cline{3-6}
\multicolumn{2}{c|}{} &  $k$NN & PPCA & mean & normal \\
\hline
\multirow{3}{*}{AUC} & PRISM III & 0.8790 & 0.8741 & 0.8709 & 0.8735\\
& Raw Linear& 0.8437 & 0.8683 & 0.8400 & 0.8141\\
& Raw Nonlinear & 0.9011 & 0.8940 & 0.8975 & 0.9223\\
\hline
\multirow{3}{*}{$J$} & PRISM III & 0.6047 & 0.6153 & 0.5788 & 0.5840 \\
& Raw Linear& 0.5479 & 0.5889 & 0.5637 & 0.5682 \\
&Raw Nonlinear & 0.6835 & 0.7144 & 0.7198 & 0.7301\\
\hline
\multirow{3}{*}{PrecRec} & PRISM III & 0.5000 & 0.5000 & 0.5172 & 0.5172\\
& Raw Linear& 0.5000 & 0.5542 & 0.5833 & 0.5833\\
& Raw Nonlinear & 0.5172& 0.5417 & 0.5000 & 0.5152\\
\hline
\multirow{3}{*}{BS} & PRISM III & 0.0693 & 0.0685 & 0.0696 & 0.0686\\
& Raw Linear& 0.0713  & 0.0689  & 0.0707  &  0.0631\\
& Raw Nonlinear & 0.0998 & 0.0783  & 0.0911 & 0.0907 \\
\hline
\end{tabular}
\end{center}
\end{table}

Though logistic regression with linear features and variables imputed with PPCA leads
to a slight improvement in terms of $J$ and the precision-recall balance over the PRISM III scores 
without imputation, the PRISM III scores calculated with the imputed data outperforms both logistic 
scores in terms of AUC and $J$. Logistic regression with linear features and variables 
imputed with mean or normal values results in the same PrecRec value as the novel score. However, 
the novel score greatly outperforms both of these raw variable models in terms of AUC and $J$. 
These discrimination results suggest the variables in the scores should be nonlinearly 
transformed in order to achieve accurate mortality prediction.  Note that the novel score without 
imputation performs better than the scores presented in Table \ref{table:rawResults} in terms of all of 
the evaluation criteria except for the Brier score of the logistic model with linear variables and imputation 
with normal values.

A close inspection of the measures of discrimination of the logistic regression with a nonlinear 
feature set in Table \ref{table:rawResults} further emphasizes the importance of performing the nonlinear 
transformation of physiological variables. Despite slightly higher AUC and $J$ achieved by the logistic 
models with cubic spline transformations of the features as compared to the novel algorithm, the novel 
algorithm outperforms the logistic models in terms of precision-recall balance and the Brier score. These two 
measures are of particular significance when considering the class imbalance and assessing accuracy
of identifying subjects at risk for in-hospital mortality. Additionally, expert knowledge is ignored in the 
logistic regression with cubic spline covariates and the interpretation of the variable contributions to the 
score is unclear since the range of increasing risk is not a byproduct of the algorithm as in the case of 
our proposed algorithm.

\subsection{Implementation and results for the novel adult population scores}\label{sec:resultsAdult}

We apply our optimization framework to design scores that preserve the clinical 
knowledge embedded in SOFA while transforming the features using nonlinear
logistic functions (as we did earlier for PRISM III and a pediatric brain injury population). 

\subsubsection{Adult patient population}
We use the data from the MIMIC II clinical database \cite{mimic2,physionet} to find the parameters of the logistic functions and test the ability of the novel score to predict mortality in the current ICU stay. The examined 
dataset consists of 3711 adult ICU patients (7.14\% mortality).

\subsubsection{Implementation of the novel prediction scheme in adult population}
The variables included in the SOFA score calculation are $\text{PaO}_2/\text{FiO}_2$, respiratory support, platelets, bilirubin, mean arterial pressure, Glasgow Coma Scale, creatinine, daily urine output, and certain cardiovascular drugs. The worst value of 
each physiological variable in the first 24 hours is used to compute SOFA and as 
a raw variable for the calculation of the new score. The features representing the 
respiratory and cardiovascular systems are treated as indicator variables since the 
respiratory SOFA calculation includes an AND statement (e.g., the respiratory SOFA=3 if 
$\text{PaO}_2/\text{FiO}_2<200$ AND the patient is receiving respiratory support) and 
because the cardiovascular SOFA greater than 1 is dependent on the administration 
of certain drugs. The urine output criterion for the renal SOFA is ignored since urine 
output was not charted consistently and thus the daily value might not be reliable. 
 The parameters of the backtracking line search and the prior on $\mathbf{w}$ 
are as specified in Section \ref{sec:implPedi} with the exception of the backtracking line 
search over $\mathbf{t}$, which was implemented with $\beta=0.9$ due to a fast
convergence of the algorithm.

\subsubsection{Performance comparison of novel score and existing score in adult population}
\begin{figure}[!h]
\begin{centering}
\includegraphics[width=5.5in]{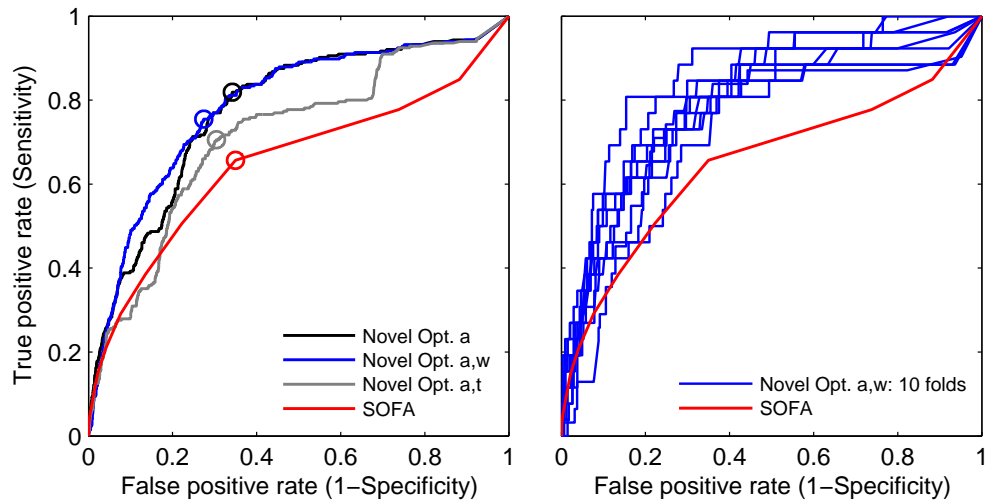}
\caption{ROC curves for SOFA (red) and the novel score preserving expert 
knowledge from SOFA. The left panel shows the ROC curves generated using the outputs 
of the 10-fold cross-validation for the optimization over nonlinear transformation slopes 
$\mathbi{a}$ (black), and alternating minimization over the slopes $\mathbi{a}$ and weights 
$\mathbf{w}$ (blue) and the slopes $\mathbi{a}$ and soft thresholds $\mathbf{t}$ (gray). The 10 ROC curves from the cross-validation folds of the novel score optimized over $\mathbi{a}$ and $\mathbf{w}$ are presented in the right panel (blue) along with the ROC curve for SOFA (red).}
\label{fig_ROCadult}
\end{centering}
\end{figure}

The novel score is compared to SOFA in terms of the same discrimination (AUC, J, 
precision-recall balance) and calibration (Brier score) criteria described in Section 
\ref{sec:testing}. The average discrimination and calibration results
 from 10-fold cross-validation for the novel scores with optimization over various parameters are presented in Table \ref{ROCtable_SOFA}. In order to visually 
compare the discriminatory capabilities of the novel score optimized over $\mathbi{a} 
\text{ and } \mathbf{w}$, we also present the ROC curve calculated from the risk scores 
obtained from the cross-validation results (in the left panel of Figure \ref{fig_ROCadult}) and 
the ROC curves from the 10 folds (in the right panel of Figure \ref{fig_ROCadult}). It should 
be noted that the novel score has higher discrimination than SOFA for every fold.

\begin{table}
%% increase table row spacing, adjust to taste
\renewcommand{\arraystretch}{1.3}
\caption{Risk Score Accuracy: Adult Population}
\label{ROCtable_SOFA}
\centering
\begin{center}
\begin{tabular}{|c|c|c|c|c|}
\hline
Score & AUC & $J$ & PrecRec & BS \\
\hline
SOFA &  0.6514 &  0.3069 & 0.2258 & 0.0630\\
\hline
Novel Score optimized over $\mathbi{a}$ & 0.7753  &0.5213&0.3188 & 0.0603\\
\hline
Novel Score optimized over $\mathbf{w}$ & 0.7921&0.5357 &0.2886&0.0666\\
\hline
Novel Score optimized over $\mathbi{a}$ and $\mathbf{t}$ &0.7167 &0.4498  & 0.2678& 0.0641\\
\hline
Novel Score optimized over $\mathbi{a}$ and $\mathbf{w}$ &  0.7885 &  0.5275 & 0.3191&0.0617\\
\hline
\end{tabular}
\end{center}
\end{table}

The novel scores outperform SOFA in all of the discrimination criteria, while the score 
optimized over the slopes of the nonlinear transformations $\mathbi{a}$ and the score 
optimized over $\mathbi{a}$ and the nonlinear feature weights $\mathbf{w}$ also yield 
better Brier scores. The highest AUC and $J$ are obtained when optimizing over the 
weights $\mathbf{w}$. Including the slopes $\mathbi{a}$ in the optimization along with 
$\mathbf{w}$ results in a slight decrease in AUC and $J$ but leads to an improvement 
of the precision-recall balance and the Brier score. This score, optimized over 
$\mathbi{a}$ and $\mathbf{w}$, might therefore be preferred given the class-imbalance 
of the problem and the importance of identifying true positives. The novel score 
optimized over the slopes $\mathbi{a}$ and soft thresholds $\mathbf{t}$ exhibits 
worse discrimination than the scores obtained via other optimizations despite achieving 
the lowest objective value on the training sets. Upon further examination, we found 
that the final score no longer retained the step-wise structure of SOFA. More 
specifically, the final soft thresholds either collapsed into a single step or were outside the 
physiological range for some variables and thus had minor contribution to the final score. 
This finding stresses the importance of preserving the clinical knowledge embedded in the 
original score; to perform well on the test set, the novel score structure and optimization 
methodology should result in a score that resembles the original one.

\section{Conclusions} \label{sec:conclusion}

We have developed a novel outcome prediction score that exploits advantages of additive 
stepwise risk scores and addresses the key limitation of hard thresholds typically used by 
state-of-the-art prediction methods. In particular, by transforming predictive variables using a 
combination of logistic functions, the developed method allows for a fine differentiation between 
critical and normal values of the predictive variables. Optimization of the continuous score 
allows for not only specifying different weights for the variables but, by optimizing over the 
slope and/or inflection point of the logistic curve in the feature transformation, we can also 
identify the range of values of each variable where the risk increases. This optimization need 
only be performed once to determine the optimal parameters and the score is thereafter quickly 
calculated for each patient. Optimal values of the parameters of logistic functions may be readily 
re-learned as the patient population and standards of care evolve. The novel scores 
derived using the proposed optimization framework demonstrate significantly higher predictive 
power than the widely used PRISM III in a pediatric brain trauma population and the SOFA in 
an adult ICU population.

The presented method can be broadly applied to devise and optimize risk scores with predictive 
power superior to schemes that use hard-thresholding of physiological variables, 
as has been shown in the cases of PRISM III and SOFA. Future applications of 
the developed scheme include optimization of in-hospital mortality risk 
scores designed for adult ICU population, such as the Acute Physiology and 
Chronic Health Evaluation (APACHE) \cite{zimmerman_2006,knaus_1991} and the Simplified 
Acute Physiology Score (SAPS) \cite{moreno_2005}. Moreover, the proposed
method may be used to develop risk prediction scores geared towards sub-populations of interest 
when a hard-thresholding score exists for a larger population (e.g. specific diseases within an 
ICU population) or in geographical areas with different standards of care than those used in the 
development of the existing scores.

\section*{Acknowledgments}
The authors would like to thank Karen Piper for the identification of patients that met the inclusion criteria for the study. This material is based upon work supported by the National Science Foundation Graduate Research Fellowship under Grant No. DGE-1110007.




%\section*{References}
\bibliographystyle{IEEEtran} 
%\bibliography{logistic.bib}

\end{document}